\documentclass[letterpaper, 10 pt, conference]{ieeeconf}
\IEEEoverridecommandlockouts    % This command is only needed if
\usepackage{graphics}           
\usepackage{times}              
\usepackage{amsmath}            
\usepackage{amssymb}            
\usepackage{graphicx}
\usepackage{algorithm}
\usepackage[noend]{algpseudocode}
\usepackage{booktabs}
\usepackage{color}
\usepackage[nocompress]{cite} % sorts citations automatically, e.g., "[1],[2],[3]" instead of "[2],[3],[1]".
\definecolor{instructioncolor}{rgb}{.5,.5,.5}

\usepackage[font=small]{caption}

\def\eqref#1{Eq.~(\ref{#1})}

\makeatletter
\usepackage{xspace}
\DeclareRobustCommand\onedot{\futurelet\@let@token\@onedot}
\def\@onedot{\ifx\@let@token.\else.\null\fi\xspace}

\makeatother

\usepackage{array}
\newcolumntype{L}[1]{>{\raggedright\let\newline\\\arraybackslash\hspace{0pt}}m{#1}}
\newcolumntype{C}[1]{>{\centering\let\newline\\\arraybackslash\hspace{0pt}}m{#1}}
\newcolumntype{R}[1]{>{\raggedleft\let\newline\\\arraybackslash\hspace{0pt}}m{#1}}

\newcommand{\RR}{\mathbb{R}}

\title{\LARGE \bf Calibration-Informative Region Selection for Online LiDAR--Camera Calibration in Agricultural Environments}

\author{Rajitha de Silva \and Grzegorz Cielniak% <-this % stops a space
  \thanks{This work was supported by
Innovate UK-funded Project Agri-OpenCore under Grant 10041179. All authors are with Lincoln Institute for Agri-Food Technology, University of Lincoln, UK. Corresponding Author: Rajitha de Silva ({\tt\small odesilva@lincoln.ac.uk})}%
}

\begin{document}
\maketitle
\thispagestyle{empty}
\pagestyle{empty}

%%%%%%%%%%%%%%%%%%%%%%%%%%%%%%%%%%%%%%%%%%%%%%%%%%%%%%%%%%%%%%%%%%%%%%%%%%%%%%%%
\begin{abstract}
  %
  %% WHY 
  % Use 1-2 not too long sentences, which clearly answer the WHY question: 
  % Why is this relevant, why should I care? Motivate why the stuff that 
  % you enable is relevant (not neccessarily equal to the technique)

  %% WHICH PROBLEM 
  % One sentence that explain the problem the paper addresses/ivestigates
  % Start with: In this paper, we address the problems of \dots

  %% HOW & WHAT
  % Around 3 sentences that explain how to approach the problem in general and answers:
  % How to solve the problem in general? (1/2 - 1 sentence)
  % What makes our approach special? What are we actually doing? What is new?
  
  %% IMPLEMENTATION, EVALUATION, WHAT FOLLOWS
  % 1-2 sentences what the experiments show and potentiall what follows from
  % your great work for the research community or the rest of the world ;-)
Reliable multi-modal calibration requires identifying which observations truly constrain the extrinsic parameters and which ones mainly add noise or ambiguity. In this paper, we propose a support-map-driven approach to multi-modal calibration that decouples four functional blocks: initial calibration, cross-modal residual extraction, support-map estimation, and support-aware refinement. We instantiate this formulation for online LiDAR--camera calibration using MDPCalib, a target-less LiDAR--camera calibration method based on motion and deep point correspondences, and CMRNext, a dense LiDAR--camera matching model that predicts optical-flow-like image-plane residuals. The key contribution is a dense \emph{calibration support map} that aggregates cross-modal agreement over aligned observations and highlights where calibration evidence is consistently reliable. Across the Bacchus Long-Term (BLT) dataset and KITTI, we show that calibration evidence is spatially and semantically non-uniform, indicating that some semantic regions provide stronger cues for calibration than others. On KITTI, support-guided refinement improves the calibration performance with better translation accuracy while rotational gains remain limited.
\end{abstract}

%%%%%%%%%%%%%%%%%%%%%%%%%%%%%%%%%%%%%%%%%%%%%%%%%%%%%%%%%%%%%%%%%%%%%%%%%%%%%%%%
\section{Introduction}
\label{sec:intro}

%%%%%%%%%%%%%%%%%%%
%% WHY: 
% First, answer the WHY question: Why is that relevant? Why should I be
% motivated to read the paper? Why should I care? (1 paragraph, 2-5 sentences)

%\textbf{Why} ...

%%%%%%%%%%%%%%%%%%%
%% WHICH PROBLEM
% Second, explain WHICH problem you are solving/address to solve.

%\textbf{Problem:} In this paper, we investigate/consider the problem of ...

%%%%%%%%%%%%%%%%%%%
%% HOW & WHAT
% Third, explain briefly how one can address the problem in general and mention 
% briefly what others/we before have done. Prepare the reader for your contribution 
% that comes in the next section (and not here!).

%textbf{How \& What:}

% Link to figure somewhere
% See \figref{fig:motivation} for an example.

%%%%%%%%%%%%%%%%%%%
%% MAIN CONTRIBUTION & WHAT FOLLOWS FROM THAT
% Explain your contribution in one paragraph. This is a very important paragraph. 
% Always start that paragraph with: ``The main contribution of this paper is''
% and be SUPER-EXPLICIT for what you claim contribution (and maybe novelty) for.

%%%%%%%%%%%%%%%%%%%
%% OUR KEY CLAIMS (can be merged with the main contribution above if desired)
% Explicitly(!) state your claims in one (short) paragraph and make
% sure you pick them up again in the experiments and support every claim.

Accurate multi-modal calibration is essential for robotic perception, as extrinsic errors propagate directly into localization, mapping, and scene understanding. In long-term field robotics, calibration can drift because of vibration, thermal variation, and minor mechanical changes, while recalibration with targets or dedicated rigs is often impractical once the platform is deployed. We focus on agriculture because it is a particularly demanding setting, where repetitive vegetation, occlusions, weak texture, and only a few rigid structures make calibration evidence highly non-uniform across the scene.

Recent target-less calibration methods provide the ingredients for online refinement. MDPCalib~\cite{mdpcalib} combines motion alignment with learned 2D--3D correspondences for automatic camera--LiDAR calibration, while CMRNext~\cite{cmrnext} provides generalizable dense camera--LiDAR matching by recasting point--pixel association as optical flow. However, they often treat correspondences from all parts of the environment as equally informative, so weak or ambiguous regions can affect the estimate as much as stable structures, reducing calibration accuracy and robustness. A similar intuition appears in long-term agricultural perception: Hroob et al.~\cite{hroob2024generalizable} show, from the 3D structural perspective, that persistent scene elements improve localization in BLT vineyard data, whereas vegetation and other changing elements are less reliable; de Silva et al.~\cite{ksi} show, from the visual-semantic perspective, that stable landmarks yield more distinctive matches than non-stable regions. Together, these observations suggest that calibration should explicitly estimate where reliable cross-modal evidence originates rather than treating all regions as equally informative.

\begin{figure}[t]
  \centering
  \includegraphics[width=0.98\linewidth]{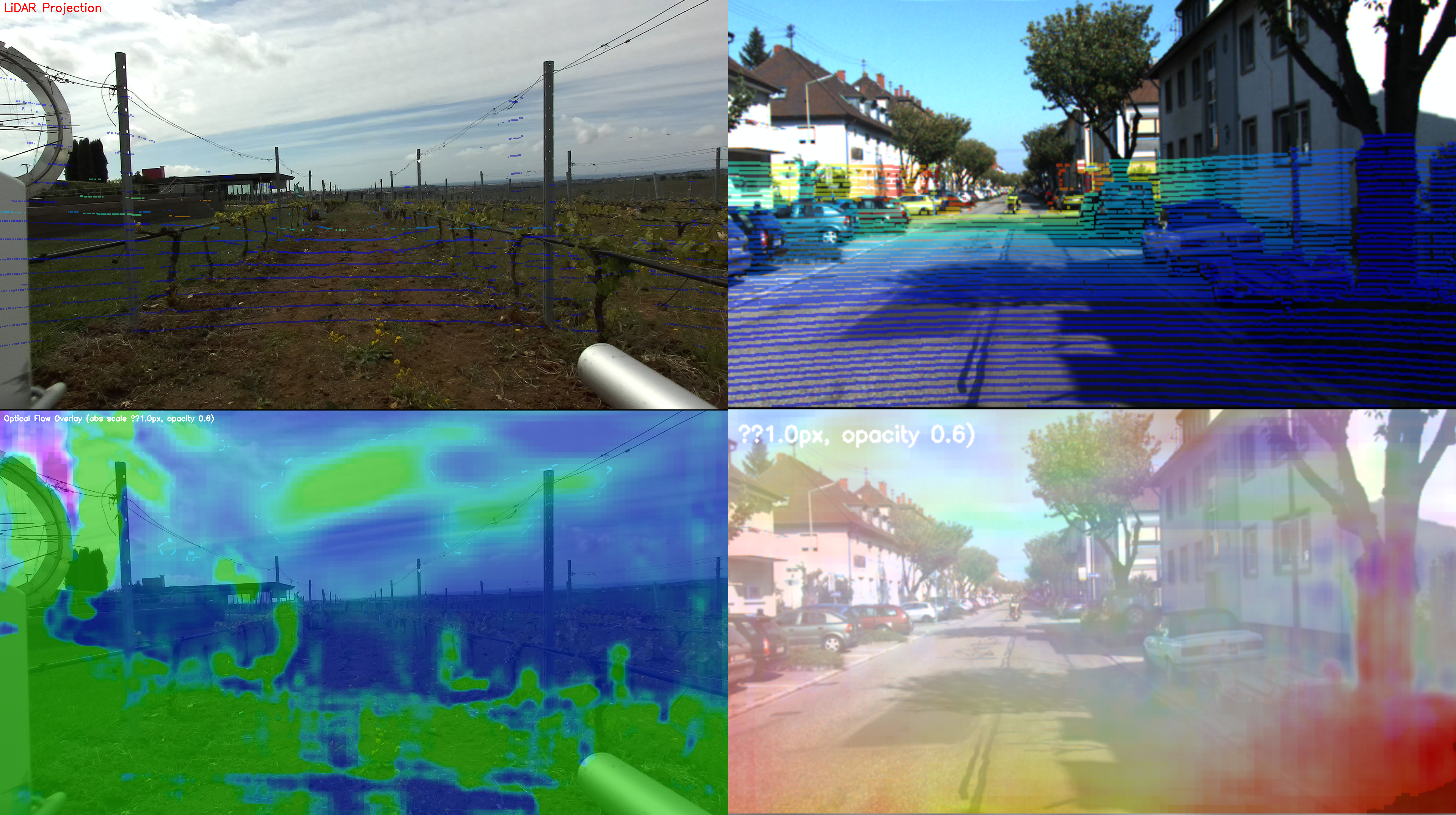}
  \caption{Representative examples from two datasets: BLT agricultural field data on the left and KITTI urban driving data on the right. The top row shows LiDAR projections under the initial extrinsic estimate, and the bottom row shows the corresponding calibration support maps.}
  \label{fig:motivation}
\end{figure}

Motivated by this, we formulate calibration as a support-estimation problem in addition to a correspondence-matching problem. Our goal is to introduce a modular calibration view in which any pipeline that provides a provisional extrinsic estimate and a cross-modal residual signal can, in principle, construct and use a calibration support map. In this paper, we instantiate that idea for online LiDAR--camera calibration. We study it primarily on the Bacchus Long-Term (BLT) dataset~\cite{blt}, a long-term agricultural dataset, and use KITTI~\cite{skitti} as an additional out-of-domain benchmark to test whether the same support structure persists beyond agriculture. Figure~\ref{fig:motivation} shows representative examples from both datasets.
%To study this, we first use the reference extrinsic calibration provided with a dataset, which is typically obtained through an external measurement or a robust calibration procedure which serves as the ground-truth alignment between the LiDAR and camera. We then analyse the calibrated image--LiDAR pairs using CMRNext~\cite{cmrnext}, a dense camera--LiDAR matching framework that predicts cross-modal agreement in an optical-flow-like manner. In this work, we use the dense agreement output of CMRNext and interpret it as a \emph{calibration support signal} that indicates which image regions are locally consistent with the current extrinsic estimate.

The main contribution of this paper is a new support-map-driven formulation of multi-modal calibration. The core idea is to represent calibration evidence as a dense spatial prior that can be estimated separately from the optimizer and then reused during refinement. In sum, we make three key claims. First, calibration evidence in agricultural LiDAR--camera observations is spatially and semantically non-uniform, with rigid scene structures providing more reliable support than foliage and other weakly constrained regions. Second, explicitly estimating this support and using it for correspondence selection improves refinement compared with uniform sampling. Third, although the experiments instantiate the method with one LiDAR--camera calibration stack and are motivated by agriculture, the support-structure intuition also appears on KITTI, indicating that the underlying idea transfers beyond a single dataset and is conceptually broader than one particular implementation.

%%%%%%%%%%%%%%%%%%%%%%%%%%%%%%%%%%%%%%%%%%%%%%%%%%%%%%%%%%%%%%%%%%%%%%%%%%%%%%%%
\section{Methodology}
\label{sec:main}
\begin{figure*}[t]
  \centering
  \includegraphics[width=0.98\textwidth]{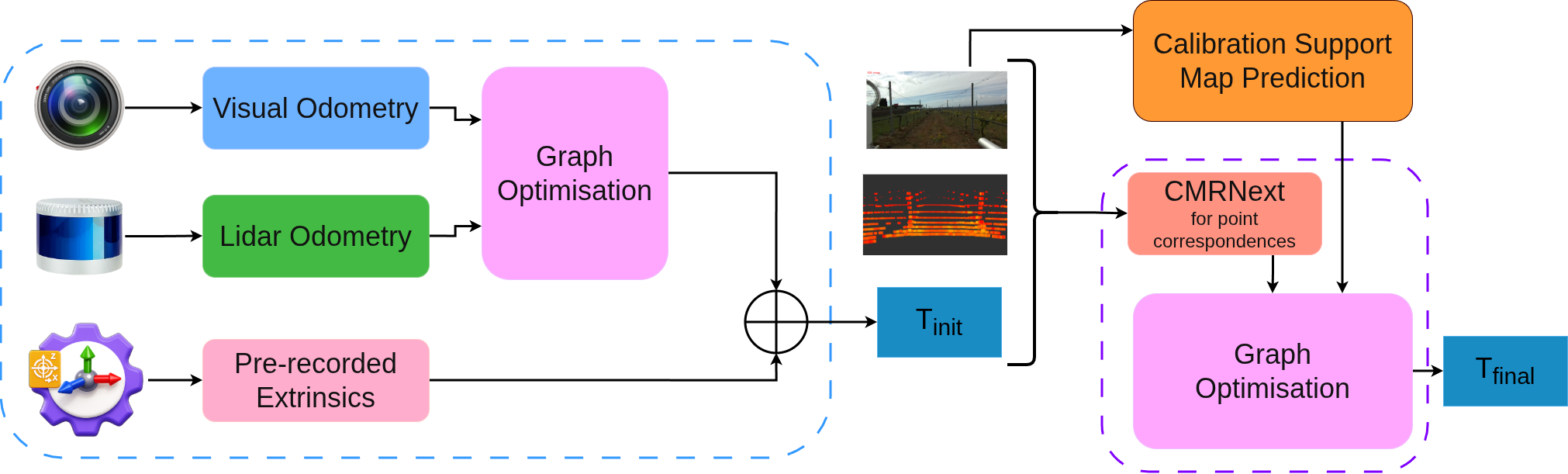}
  \caption{Support-map-driven calibration as a four-block functional pipeline. Stage~1 provides an initial extrinsic estimate $\mathbf{T}^{0}_{LC}$, Stage~2 extracts cross-modal residuals, Stage~3 converts them into a dense support map $s(\mathbf{x})$, and Stage~4 uses that map for support-aware refinement. In this paper, Stages~1 and~4 are instantiated with MDPCalib and Stage~2 with CMRNext.}
  \label{fig:pipeline}
\end{figure*}

We decompose support-map-driven calibration into four functional blocks: (1) initial calibration, (2) cross-modal residual extraction, (3) support-map estimation, and (4) support-aware refinement. As shown in Fig.~\ref{fig:pipeline}, this decomposition separates where calibration evidence occurs from how the extrinsics are optimized. In the present implementation, Stages~1 and~4 use MDPCalib and Stage~2 uses CMRNext, but other modules could be substituted as long as they provide an extrinsic estimate and cross-modal residuals. Throughout this section, $\mathbf{p}_i$ denotes a LiDAR point, $\mathbf{T}^{0}_{LC}$ the current LiDAR-to-camera estimate, $\pi(\cdot)$ the camera projection, $\mathbf{u}_i=\pi(\mathbf{T}^{0}_{LC}\mathbf{p}_i)$ the projected pixel, $\mathbf{f}_i=(\Delta u_i,\Delta v_i)$ the residual vector from the matching stage, and $\tilde{\mathbf{u}}_i=\mathbf{u}_i+\mathbf{f}_i$ the matched image location. We denote the support map by $s:\Omega\rightarrow\RR_{\ge 0}$ and write $s_i=s(\mathbf{u}_i)$ for the support value at correspondence $i$.

\subsection{Stage 1: Initial Calibration Estimate}
The first block provides the initial estimate $\mathbf{T}^{0}_{LC}$ used by the later stages. In our implementation, MDPCalib~\cite{mdpcalib} computes it by time-synchronizing visual and LiDAR odometry trajectories and solving a scaled hand--eye alignment with robust graph optimization. More generally, $\mathbf{T}^{0}_{LC}$ could also come from factory calibration, a previous online estimate, or another target-less calibration method.

\subsection{Stage 2: Cross-Modal Residual Extraction}
The second block extracts residuals between projected LiDAR points and cross-modal matches. In our implementation, CMRNext~\cite{cmrnext} predicts an optical-flow-like displacement field between the projected LiDAR image and the RGB image. For each LiDAR point $\mathbf{p}_i$, we project it with the current estimate and then apply the predicted residual to obtain the matched image location:
\begin{equation}
\mathbf{u}_i=\pi(\mathbf{T}^{0}_{LC}\mathbf{p}_i), \qquad \tilde{\mathbf{u}}_i=\mathbf{u}_i+\mathbf{f}_i.
\end{equation}
The formulation only requires that this stage output correspondences or residual vectors in a common image plane; it is not tied to CMRNext specifically.

\subsection{Stage 3: Support-Map Estimation}
Stage~3 is the key contribution of the paper. Its goal is to estimate a dense spatial prior that answers the following question: \emph{if a correspondence lands at image location $\mathbf{x}$, how consistently does that region support a reference calibration?} We construct this map offline from a reference extrinsic calibration $\mathbf{T}^{\mathrm{ref}}_{LC}$. For each synchronized image--LiDAR pair at time $t$, we project every LiDAR point with $\mathbf{T}^{\mathrm{ref}}_{LC}$ to obtain $\mathbf{u}_i^t=\pi(\mathbf{T}^{\mathrm{ref}}_{LC}\mathbf{p}_i^t)$ and run CMRNext on the aligned pair to obtain a residual vector $\mathbf{f}_i^t$. We then convert the residual magnitude into a nonnegative score $a_i^t$, with smaller residuals receiving larger scores.

The dense support map is obtained by accumulating these scores in the image plane:
\begin{equation}
s(\mathbf{x}) = \sum_t \sum_i a_i^t \exp\!\left(-\frac{\|\mathbf{x}-\mathbf{u}_i^t\|_2^2}{2\sigma^2}\right),
\end{equation}
where $\sigma$ controls spatial smoothing. After accumulation, we normalize $s(\mathbf{x})$ to the range $[0,1]$. Regions that repeatedly produce accurate cross-modal matches under the reference calibration receive high support, whereas unstable or weakly constrained regions receive low support. Thus, Stage~3 aggregates evidence across many aligned observations rather than acting as a single-frame confidence heuristic. Evaluating the map at a projected correspondence yields the scalar support value $s_i=s(\mathbf{u}_i)$ used in Stage~4.

\subsection{Stage 4: Support-Aware Refinement}
The fourth block performs the final refinement using the support information from Stage~3. Using the correspondences from Stage~2 and the support values from Stage~3, we construct a support-proportional importance distribution over correspondences,

\begin{equation}
p_i=\frac{s_i}{\sum_j s_j},
\end{equation}

and perform support-guided importance sampling (SGIS), i.e., random sampling according to $p_i$ instead of uniformly. This biases the selected set toward more informative regions and reduces the influence of weakly constrained correspondences. In our LiDAR--camera implementation, the sampled correspondences are then used in a weighted MDPCalib-style geometric objective,
\begin{equation}
\Delta\mathbf{T}^{\star}=\arg\min_{\Delta\mathbf{T}}\sum_{i\in\mathcal{K}} s_i\left\|\tilde{\mathbf{u}}_i-\pi\left(\Delta\mathbf{T}\,\mathbf{T}^{0}_{LC}\mathbf{p}_i\right)\right\|_2^2,
\end{equation}
where $\mathcal{K}$ is the sampled index set. The refined estimate is $\mathbf{T}^{\star}_{LC}=\Delta\mathbf{T}^{\star}\mathbf{T}^{0}_{LC}$. More generally, any refinement method that can use weighted correspondences could consume the same support map.

%% Describe your approach. It is okay to divide the main section
%%  into a few subsections (e.g., 2-4 subsections).

%%%%%%%%%%%%%%%%%%%%%%%%%%%%%%%%%%%%%%%%%%%%%%%%%%%%%%%%%%%%%%%%%%%%%%%%%%%%%%%%
\section{Experimental Evaluation}
\label{sec:exp}

%% Repeat the main focus/objective with one single(!) sentence starting with:
%
This section first analyzes where calibration support appears in BLT and KITTI, and then evaluates whether support-guided refinement improves calibration in the current LiDAR--camera instantiation.

%% If needed (and only then!) say also a few words about the experimental
%% setup, the datasets, and used parameters. You can use a separate subsection if you
%% want to put the focus on that but often that is not needed.}

%% Note 1: It MUST be always crystal clear (a) WHY an experiment is there
%% (e.g., to support a claim, to show that the approaches useful for real-word
%% systems, to show the performance, or to provide a baseline comparison), (b)
%% WHAT it wants to show (which claim/property exactly), and (c) HOW it aims at 
%% showing this. This is ESSENTIAL for a good evaluation. Think about when BEFORE
%% designing an experiment.  IMPORTANT: Every experiment MUST start with something 
%% like:  The next experiment is presented to show \dots and thus for supporting our 
%% first claim.

%% Note 2: Start with the most important/impressive experiment first. Make
%% his a key story of the paper. Keep the order of the claims, i.e., re-order
%% claims in the intro/before if needed. 

%%%%%%%%%%%%%%%%%%%%%%%%
\subsection{Experiment 1: BLT Calibration Agreement Analysis}
This experiment quantifies class-wise calibration agreement in BLT scenes, clarifies what the support map captures, and supports our first claim in the target agricultural setting.

We run CMRNext over the BLT sequences and summarize the class-wise mean absolute residual components $|\Delta u|$ and $|\Delta v|$.

\begin{figure}[t]
  \centering
  \includegraphics[width=0.98\columnwidth]{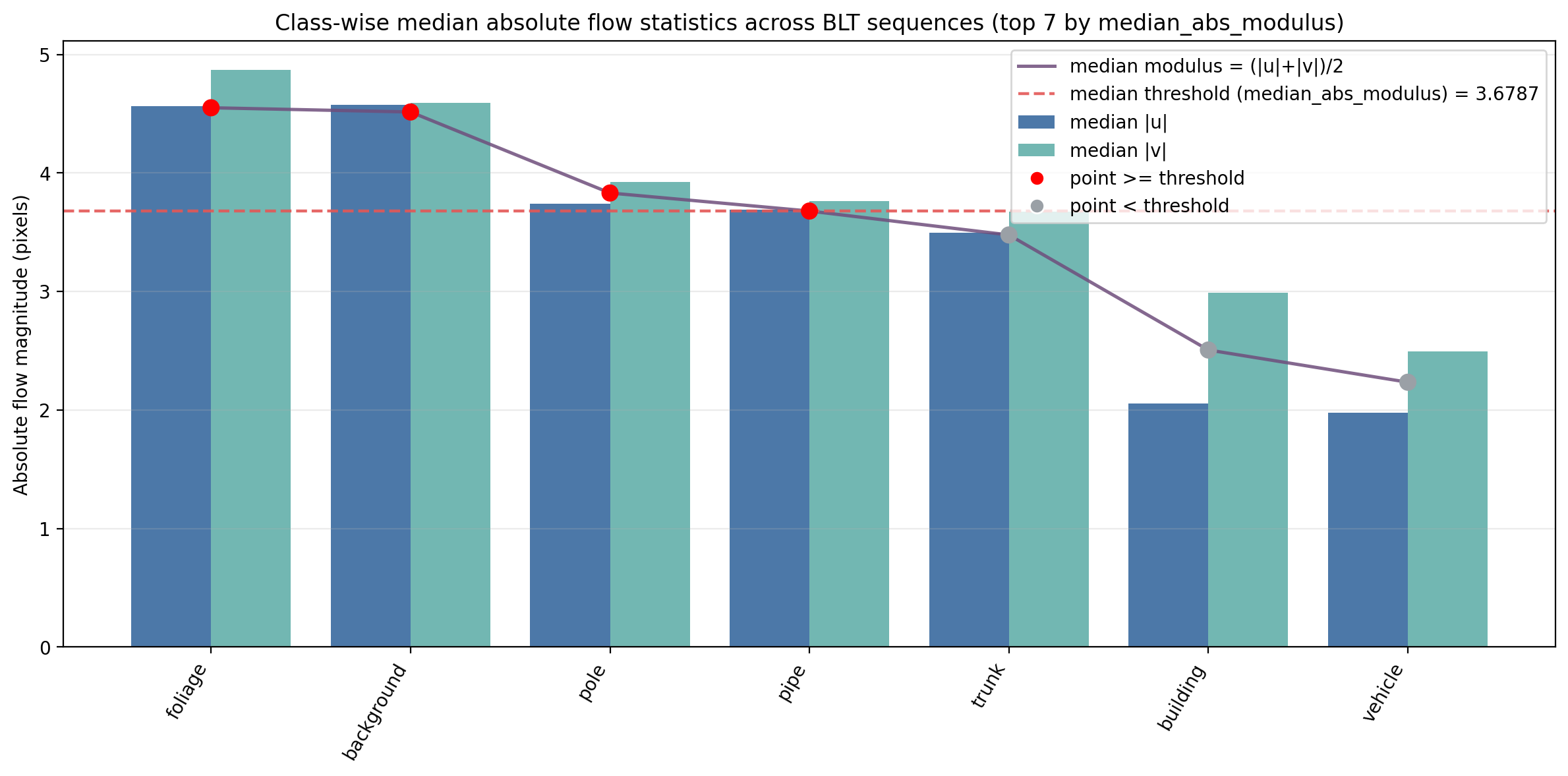}
  \caption{Experiment 1 on BLT: class-wise median calibration residual analysis using CMRNext correspondences. As in Experiment~2, the bars show median residuals in the horizontal ($\Delta u$) and vertical ($\Delta v$) directions per class, red indicates classes above the global class-wise average residual, gray indicates classes below average, and the line plot reports the per-class median summary value.}
  \label{fig:exp3-blt-semantics}
\end{figure}

The results in Fig.~\ref{fig:exp3-blt-semantics} again show strong class dependence of calibration agreement in agricultural scenes. We extract foliage using the Excess Green Index with Otsu thresholding, define background as the region not assigned to any semantic class, and obtain the remaining classes from a standard semantic segmentation model trained on the SemanticBLT~\cite{ksi} dataset. The vegetation and background dominated regions remain less supportive, whereas rigid agricultural structures such as trunks, pipes, buildings, and stationary vehicles provide more stable support. These findings motivate support-guided refinement toward geometrically stable regions in agricultural environments.

%%%%%%%%%%%%%%%%%%%%%%%%
\subsection{Experiment 2: KITTI Calibration Agreement Analysis}
\begin{figure*}[t]
  \centering
  \includegraphics[width=1.98\columnwidth]{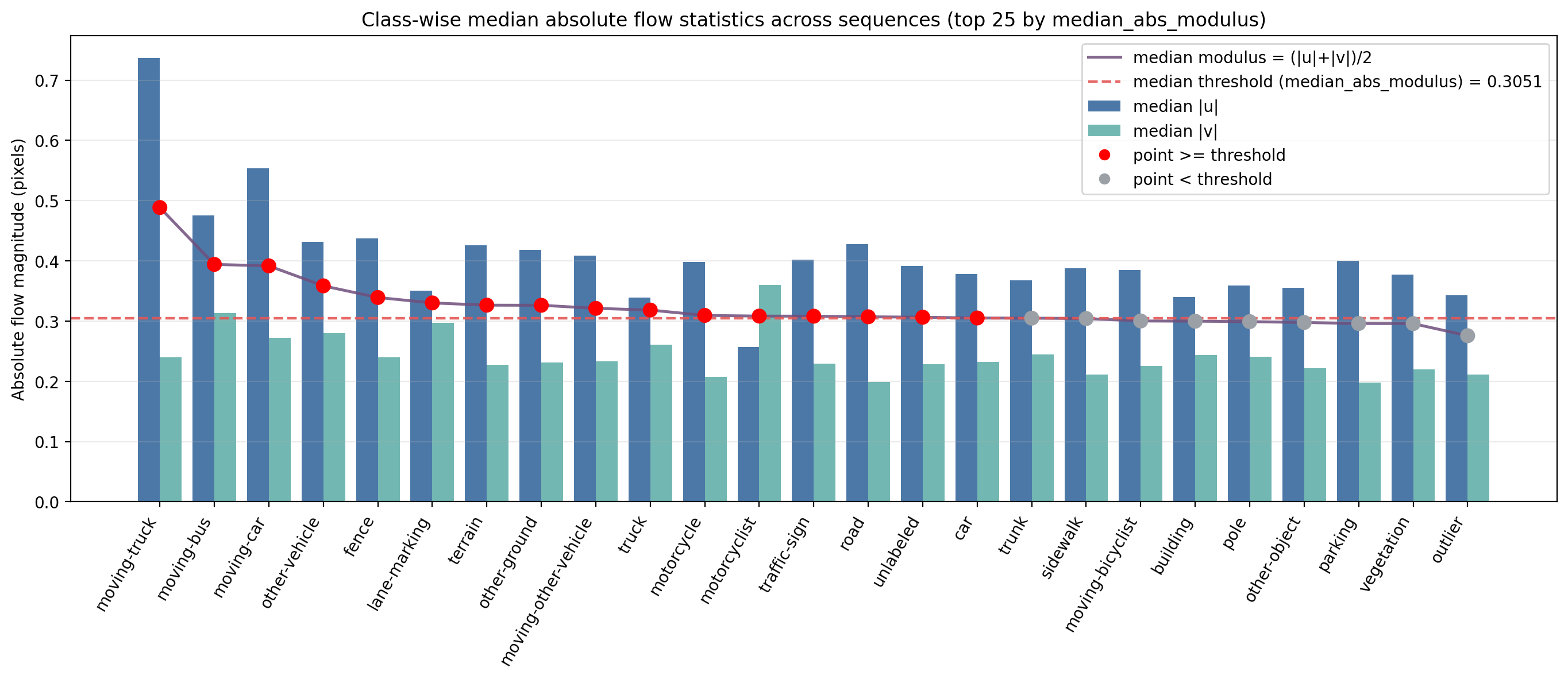}
  \caption{Experiment 2 on KITTI (sequences 00--10): the same class-wise median calibration residual analysis as in Experiment~1, now applied to KITTI.}
  \label{fig:exp1-kitti-semantics}
\end{figure*}

We repeat the same median class-wise calibration analysis for semantic KITTI using its groundn truth to test whether the same support structure appears beyond agriculture and thereby support our third claim. Figure~\ref{fig:exp1-kitti-semantics} summarizes the results for the first 11 KITTI odometry sequences (00--10).

The results in Fig.~\ref{fig:exp1-kitti-semantics} show clear semantic dependence of calibration agreement on KITTI as well. Dynamic vehicle categories dominate the high-residual end of the ranking, and classes such as fence, lane marking, terrain, and traffic sign also remain above the global class-wise average. In contrast, classes such as trunk, building, pole, vegetation, and sidewalk cluster around yet below the threshold.

Although KITTI is not the target domain of this paper, these patterns mirror the BLT observation that calibration evidence is concentrated in specific scene structures rather than uniformly distributed across the image. This supports our third claim and shows that the core idea is not tied to a single agricultural dataset. However, the class ordering differs between BLT and KITTI, suggesting that the support map reflects a combination of geometry, texture, dynamics, and dataset-specific visibility rather than a fixed semantic hierarchy. In addition, the residual magnitudes on KITTI are roughly an order of magnitude smaller than on BLT, indicating that online calibration is more challenging in agricultural scenes than in urban environments.

\subsection{Experiment 3: KITTI Support-Guided Calibration Improvement}
The final experiment evaluates whether SGIS improves final calibration on KITTI and thereby supports our second claim. Although the paper is motivated by agriculture, KITTI provides a useful external benchmark for testing whether the refinement mechanism remains beneficial beyond BLT. We compare two calibration-refinement variants over 10 evaluation runs: (i)~the original MDPCalib refinement with uniform correspondence sampling, and (ii)~the proposed support-guided refinement using SGIS derived from the calibration support map. We report final translation and rotation errors for each run. We do not report an analogous BLT refinement experiment as a positive result, because our preliminary BLT runs deviated from the reference calibration rather than improving it.

Across the 10 runs, SGIS reduces the mean translation error from $0.3171\,\mathrm{cm}$ to $0.2615\,\mathrm{cm}$ (about $17.5\%$ improvement) and the median translation error from $0.3281\,\mathrm{cm}$ to $0.2816\,\mathrm{cm}$ (about $14.2\%$ improvement). Translation stability also improves, with standard deviation decreasing from $0.1088\,\mathrm{cm}$ to $0.0718\,\mathrm{cm}$. For rotation, the mean changes from $0.04881^{\circ}$ to $0.04996^{\circ}$, while the median remains nearly unchanged ($0.04854^{\circ}$ vs. $0.04868^{\circ}$) and the standard deviation decreases from $0.00478^{\circ}$ to $0.00451^{\circ}$. At run level, SGIS improves translation in 7/10 runs and rotation in 5/10 runs.

Overall, these results indicate that support-guided refinement provides more accurate and stable translational calibration while maintaining only comparable rotational performance. The gains are concentrated in translation, and the current pipeline still relies on support maps generated offline. However, this limitation could be addressed by training a dense image-based predictor of the support map using the currently generated offline maps as surrogate ground-truth targets. The added support-estimation stage, however, also introduces computational overhead and slows calibration. We also attempted a BLT analogue of Experiment~3, but the refined calibration deviated from the reference calibration. The visual and LiDAR odometry estimates used by the current pipeline may be responsible: in agricultural environments, repetitive vegetation and weak structure can make these motion estimates error-prone and impose unreliable constraints on calibration.
%%%%%%%%%%%%%%%%%%%%%%%%%%%%%%%%%%%%%%%%%%%%%%%%%%%%%%%%%%%%%%%%%%%%%%%%%%%%%%%%
\section{Conclusions and Future Work}
\label{sec:conclusion}

In this paper, we presented a support-map-driven formulation of multi-modal calibration and instantiated it for online LiDAR--camera calibration.
Our approach separates initial estimation, residual extraction, support-map estimation, and support-aware refinement, thereby decoupling the central idea from any single calibration backbone.
Our primary results on BLT show that calibration support is strongly class-dependent in agricultural scenes: vegetation- and background-dominated regions exhibit larger disagreement, whereas rigid structures such as trunks, pipes, buildings, and stationary vehicles provide more stable support. KITTI provides additional evidence that the same support structure persists beyond agriculture and that SGIS improves translational calibration accuracy and stability on an external dataset while leaving rotational performance largely unchanged.
Overall, these findings indicate that support-map-driven correspondence selection is a promising strategy for online LiDAR--camera calibration in agricultural field robotics, while also suggesting a broader route toward support-aware multi-modal calibration.

In the current implementation, Stage~3 uses offline support maps from a previous MDPCalib+CMRNext run; replacing this stage with a dense RGB-image predictor trained from these maps is an important next step. Future work will prioritize agricultural validation by improving BLT odometry, adapting CMRNext to field imagery, and reassessing support-guided refinement with independent calibration references in real deployments.

\section*{Acknowledgments}
Generative AI tools (including ChatGPT, Gemini, and Claude) assisted with visualization assets, manuscript refining, and code development. The authors reviewed, edited, and verified all AI-assisted content and take full responsibility for the manuscript and artifacts.
%%%%%%%%%%%%%%%%%%%%%%%%%%%%%%%%%%%%%%%%%%%%%%%%%%%%%%%%%%%%%%%%%%%%%%%%%%%%%%%%
%% Future work: Use only if applicable -- but if so, use the following
%% sentence to start:
% Despite these encouraging results, there is further space for improvements. 
%% In general, I avoid explaining future work in 6-8 page conference papers...

%%%%%%%%%%%%%%%%%%%%%%%%%%%%%%%%%%%%%%%%%%%%%%%%%%%%%%%%%%%%%%%%%%%%%%%%%%%%%%%%
% Only if applicable
%\section*{Acknowledgments}
%We thank XXX for fruitful discussions and for \dots

\bibliographystyle{ieeetr}

% All new citations should go to new.bib. The file glorified.bib should go
% be the one from the ipb server. After paper or related work has been
% written merge the entries from new.bib to glorified.bib ON THE SERVER,
% replace the glorified.bib in this repository and empty the new.bib
\bibliography{glorified,new}

\end{document}